\documentclass{article}
\usepackage{spconf,amsmath,graphicx}
\usepackage[utf8]{inputenc} 
\usepackage[T1]{fontenc}    
\usepackage{hyperref}       
\usepackage{url}            
\usepackage{booktabs}       
\usepackage{amsfonts}       
\usepackage{nicefrac}       
\usepackage{microtype}      
\usepackage{xcolor}         
\usepackage{bm}
\usepackage{amssymb}
\usepackage{setspace}
\usepackage{adjustbox}

\makeatletter

\newcommand{\parn}[1]{\left(#1\right)}

\newlength{\NOTskip}

\title{Phoneme-Level BERT for Enhanced Prosody of Text-to-Speech with Grapheme Predictions}
\name{Yinghao Aaron Li, Cong Han, Xilin Jiang, Nima Mesgarani}
\address{Department of Electrical Engineering, Columbia University, USA}
\begin{document}
%
\maketitle
\begin{abstract}
  Large-scale pre-trained language models have been shown to be helpful in improving the naturalness of text-to-speech (TTS) models by enabling them to produce more naturalistic prosodic patterns. However, these models are usually word-level or sup-phoneme-level and jointly trained with phonemes, making them inefficient for the downstream TTS task where only phonemes are needed. In this work, we propose a phoneme-level BERT (PL-BERT) with a pretext task of predicting the corresponding graphemes along with the regular masked phoneme predictions. Subjective evaluations show that our phoneme-level BERT encoder has significantly improved the mean opinion scores (MOS) of rated naturalness of synthesized speech compared with the state-of-the-art (SOTA) StyleTTS baseline on out-of-distribution (OOD) texts.  
\end{abstract}
\begin{figure*}[!th]
    \centering
    \includegraphics{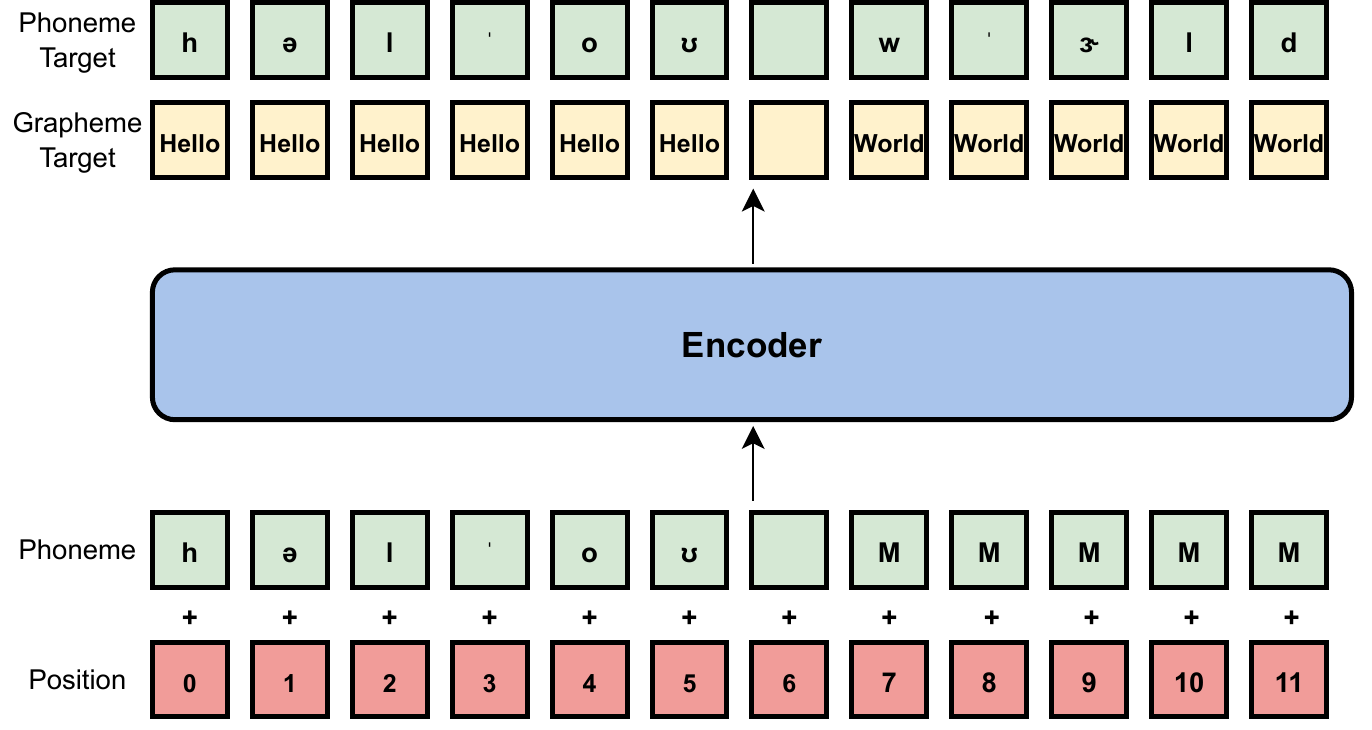}
    \caption{Pre-training scheme for phoneme-level BERT. $M$ indicates that the input token has been masked. The transformer encoder takes the phoneme tokens and their position encoding as the input and is trained to predict the masked phoneme tokens and their corresponding grapheme tokens. }
    \label{fig:1}
\end{figure*}
\section{Introduction}
\label{sec:intro}
Text-to-speech (TTS) has seen significant progress in recent years, and the most recent works are shown to synthesize speech indistinguishable from natural human speech for in-distribution texts evaluated subjectively by human raters \cite{tan2022naturalspeech}. Despite many recent advancements, it remains a challenge to synthesize natural and expressive speech due to the rich information contained in the prosody and emotions of human speech \cite{tan2021survey}. One crucial aspect that is difficult to capture in many TTS models is the tone, or the prosody, of speech \cite{li2022styletts}. Training a TTS model is like learning a language from scratch. It is crucial to have hundreds of hours of input to learn the correct intonations and emotions of a foreign language. Even with these many hours of input, non-native speakers can still be easily recognized from their intonations and prosodies. TTS datasets, on the other hand, usually contain far less data than hundreds of hours due to the requirement of data annotation. With merely a few hours of data, it is expected that the trained models will have difficulties capturing naturalistic prosodic patterns with plain phonemes as input. Hence, large-scale pre-trained models are needed to alleviate this problem. BERT, in particular, has proven effective in improving the performance of TTS models at either word level\cite{kenter2020improving}, character level \cite{xiao2020improving, zhang2020unified}, or sentence level \cite{xu2021improving}. 

Despite its success in improving the prosody and naturalness of speech synthesis, these BERT models are not trained at the phoneme level, even though the input to the downstream TTS task consists of the phonemes only. PnG-BERT \cite{jia2021png} has attempted to tackle this problem by jointly training with phoneme and grapheme tokens as input and predicting masked tokens for both phonemes and graphemes. This approach learns richer representations from both graphemes and phonemes, but it only works for a fixed set of tokens of graphemes and can fail for unseen words during training. In addition, the number of tokens for graphemes is prohibitively large, making the model slow for training and inference. A recent work, Mixed-Phoneme BERT (MP-BERT) \cite{zhang2022mixed}, leverages the need for graphemes by training a BERT model that only takes phonemes as the input. Since the phonemes are not as linguistically expressive as the graphemes, MP-BERT also learns a set of sup-phoneme units using the byte-pair encoding (BPE) \cite{sennrich2015neural} that enhances the semantic content of learned representations. MP-BERT demonstrates performance comparable to PnG-BERT for downstream TTS tasks, albeit no grapheme input is required. However, there is no guarantee that the sup-phoneme units learned through BPE carry as much linguistic information as graphemes. In addition, the number of tokens needed for sup-phoneme units is as large as 30,000 in \cite{zhang2022mixed}, greatly limiting the speed of training and inference. 

Here, we propose a phoneme-level BERT model for text-to-speech synthesis. By combining whole-word masked phoneme and grapheme predictions, we obtain a phoneme-level language model that is more efficient than MP-BERT without needing graphemes or sup-phoneme units as input. Our contribution lies in the additional pretext task that predicts the corresponding graphemes for each phoneme (phoneme-to-grapheme, P2G). By learning a language model directly at the phoneme level, the model produces representations with a deep grasp of the dynamics between phonemes, words, and semantics, therefore improving the performance of downstream TTS tasks. Subjective evaluations show that our phoneme-level BERT significantly outperforms the current state-of-the-art baseline StyleTTS model \cite{li2022styletts} in terms of perceived naturalness of speech for out-of-distribution (OOD) texts. We also demonstrate that evaluations for in-distribution texts are not as effective as OOD texts and propose a future direction for TTS research that shrinks the gap in performance between in-distribution and OOD texts. The audio samples can be listened to at \url{https://pl-bert.github.io/}.

\section{Phoneme-Level BERT}
\label{sec:method}
\subsection{Phoneme Representation} In phoneme-level BERT, we only take phonemes as the input because phonemes are the only information needed for intelligible speech synthesis. We do not use grapheme or sup-phoneme representations as those used in \cite{jia2021png} and \cite{zhang2022mixed} because of the enormous vocabulary size that slows down both training and inference. In addition, extra representations beyond phonemes suffer from out-of-vocabulary problems where unseen words or sup-phoneme units can occur during inference. Using phoneme-only representation solves these problems, making the pre-trained encoder an immediate drop-in replacement for text encoders of any TTS system. 

\subsection{Pre-training} Similar to the original BERT, phoneme-level BERT can be trained in a self-supervised manner on any corpus where phonemes and graphemes can be obtained in pairs. The phonemes and their corresponding graphemes can be prepared using an external grapheme-to-phoneme (G2P) tool. The graphemes can range from characters to sub-word units to whole words. Further grapheme-phoneme alignments through a dynamic programming algorithm may be required because pronunciations of a character or sub-word unit can change depending on the context in many languages, such as Japanese. For simplicity, we only use the whole words as tokens to avoid additional grapheme-phoneme alignment. 

\subsubsection{Training Objectives}

There are two objectives for pre-training: masked phoneme token prediction (MLM) and phoneme-to-grapheme (P2G) prediction. As in the original BERT model, we predict the masked input phoneme tokens from the hidden states of the last layer using a linear projection along with a softmax function. The loss function is the cross-entropy loss commonly used for multi-class prediction. For each phoneme token, we also map its hidden state to predict its corresponding grapheme with the same procedure. We calculate the MLM loss values only for the masked tokens, while we calculate the loss of P2G for all input tokens. As we show in section \ref{sec:abla}, this objective is important to learn a meaningful phoneme-level language representation for significant improvement in TTS tasks. The training objectives can be written as follows:

\begin{equation} \label{eq1}
\mathcal{L}_{MLM} = \mathbb{E}_{\bm{x}, I, \bm{y}_p}\left[\sum\limits_{i \in I}{\textbf{CE}(P_{MLM}\parn{E(\bm{x})}_i, {\bm{y_p}}_i)}\right], 
\end{equation}

\begin{equation} \label{eq2}
\mathcal{L}_{P2G} = \mathbb{E}_{\bm{x}, \bm{y}_g}\left[\sum\limits_{i=1}^N{\textbf{CE}(P_{P2G}\parn{E(\bm{x})}_i, {\bm{y_g}}_i)}\right], 
\end{equation}

\noindent where $\bm{x}$ is the masked input phoneme tokens, $\bm{y}_p$ is the original unmasked phoneme token labels, $\bm{y}_g$ is the corresponding grapheme token labels, $I$ is the masked indices, $E$ is our phoneme-level BERT encoder, $P_{MLM}$ is the linear projection for the MLM task, $P_{P2G}$ is the linear projection for the P2G task, $N$ is the total length of the text, and $\textbf{CE}(\cdot)$ denotes the cross-entropy loss function. 




\subsubsection{Masking Strategy} Since our goal is to learn a phoneme-level language model, we need to mask at the word level for the model to learn meaningful semantic representations. This masking strategy is termed whole-word masking \cite{cui2021pre} and is shown to be the most effective masking strategy for BERT models that take phonemes as input \cite{jia2021png, zhang2022mixed}. We employ the whole word masking and follow previous works \cite{jia2021png, zhang2022mixed, liu2019roberta} where the phoneme tokens of 15\% of graphemes in each sequence are selected to be masked at random. When a grapheme is selected, its phonemes tokens are replaced with a \verb|MSK| token 80\% of the time, are replaced with random phonemes token 10\% of the time, and stay unchanged 10\% of the time.

\section{Experiments} 
\subsection{Datasets}
\subsubsection{Text Pre-training Data}

We pre-train our phoneme-level BERT model on the English Wikipedia corpus consisting of 6,280,802 articles and approximately 74M sentences. We divide the dataset into a split where 6M articles are used for training, 140k articles are used for validation, and the rest are used for testing. The texts are normalized to match the pronunciations for each word using NeMo \cite{kuchaiev2019nemo}. Phonemes are obtained using Phonemizer \cite{Bernard2021} that converts text sequences into the International Phonetic Alphabets (IPA) with the eSpeak backend. 

\subsubsection{TTS Fine-tuning Data}
We use the LJSpeech dataset \cite{ljspeech17} to evaluate the performance of the downstream TTS tasks. The LJSpeech dataset consists of 13,100 short audio clips with a total duration of approximately 24 hours. The dataset is divided into a split where the training set contains 12,500 samples, validation set 100 samples and test set 500 samples. We extract mel-spectrograms with a FFT size of 2048, hop size of 300, and window length of 1200 in 80 mel bins. We synthesized waveforms from mel-spectrograms using Hifi-GAN vocoder \cite{kong2020hifi}.

\subsection{Training Details} Our phoneme-level BERT  is a 12-layer ALBERT \cite{lan2019albert} model with a hidden size of 768, an intermediate size of 2,048, and 12 attention heads. The training was conducted on 3 Nvidia A40 GPUs with a maximum length of 512 tokens and a batch size of 192 samples. For the MP-BERT baseline, we used the BPE base dictionary of 30,000 sup-phoneme units as in \cite{zhang2022mixed}. The models were trained for 1M steps, roughly 10 epochs. 

We fine-tuned our PL-BERT at the second stage of training of StyleTTS for 100 epochs. We froze the weights of PL-BERT for the first 50 epochs and fine-tuned it for another 50 epochs to make the training more stable. 

\begin{table}[!t]
	\centering
	\caption{Comparison of evaluated MOS with 95\% confidence intervals on LJSpeech dataset and CMOS with PL-BERT. \\} 
    \begin{tabular}{c|c|c}
    \hline
    Model & MOS & CMOS \\
    \hline
    Ground Truth      & 4.34 ($\pm$ 0.09) &  0.25 \\
    StyleTTS w/ PL-BERT      & {4.24} ($\pm$ 0.10) &  0.00  \\
    StyleTTS     & 4.19 ($\pm$ 0.10) & -- 0.02 \\
    
    \hline
    \end{tabular}
    \label{tab:t1}
\end{table}

\begin{table}[!t]
	\centering
	\caption{Comparison of MOS with 95\% confidence intervals (CI) and comparative MOS (CMOS) with StyleTTS w/ MP-BERT in the out-of-distribution (OOD) set. We include MOS (LJ) of in-distribution texts from \cite{li2022styletts} as a comparison. \\} 
 \begin{adjustbox}{max width=\columnwidth}

    \begin{tabular}{c|c|c|c}
    \hline
    Model & MOS (LJ) \cite{li2022styletts} & MOS (OOD) & CMOS \\
    \hline
    StyleTTS w/ PL-BERT & -- & \textbf{3.64} ($\pm$ \textbf{0.09}) & \textbf{0.16}  \\
    StyleTTS w/ MP-BERT & -- & 3.55 ($\pm$ 0.10) & 0.00 \\
    StyleTTS  & \textbf{4.01}  ($\pm$ \textbf{0.05}) & {3.49}  ($\pm$ {0.09}) & -- 0.39 \\
    VITS & 3.78 ($\pm$ 0.06) & 3.37 ($\pm 0.10$) & --- \\
    FastSpeech 2 & 2.97 ($\pm$ 0.06) & 2.86 ($\pm 0.11$) & ---  \\
    Tacotron 2 & 3.01 ($\pm$ 0.06) & 2.67 ($\pm 0.11$) & ---  \\
    \hline
    \end{tabular}
    \end{adjustbox}

    \label{tab:t2}
\end{table}

\subsection{Evaluations} We performed subjective evaluations on the mean opinion score of naturalness (MOS) to measure the naturalness of synthesized speech. We recruited native English speakers located in the U.S. to participate in the evaluations on Amazon Mechanical Turk. In every experiment, we randomly selected 30 sentences from the test set of both LJSpeech dataset (in-distribution) and Gutenberg books dataset \cite{gerlach2020standardized} (out-of-distribution). The latter is considered out-of-distribution (OOD) because the books used for testing have never been seen during training for both the pre-trained BERT model and the fine-tuned TTS model. On the contrary, since the LJSpeech dataset consists of seven audiobooks, the books to which the texts in the test set belong are already seen during training. 

For each text, we synthesized speech using StyleTTS fine-tuned with our phoneme-level BERT model, StyleTTS fine-tuned with MB-BERT, and the baseline StyleTTS model without BERT. The reference audios were selected from the training set with the highest sentence embedding similarity computed using sentence-BERT \cite{reimers-2020-multilingual-sentence-bert} between the training texts and the target text. Each speech set was rated by ten raters on a scale from 1 to 5 with 0.5-point increments. When evaluating each set, we randomly permuted the order of the models and instructed the subjects to listen and rate them without telling them the model labels \cite{li2021starganv2, li2022stylettsvc}.
We included distorted speech as the attention checker and all ratings were dropped from our analysis if the distorted speech was not rated the lowest. In addition to StyleTTS, we have also included  Tacorton 2 \cite{shen2018natural}, FastSpeech 2 \cite{ren2021fastspeech},
 and VITS \cite{kim2021conditional} as baseline models for comparison among OOD texts.  To check whether our PL-BERT model is helpful and its performance is better than MP-BERT, we also conducted several comparative MOS (CMOS) studies where the raters were asked to listen to only two samples and rate whether the second one was better or worse than the first one. The orders of the samples were shuffled, and the scores were set on a scale from -6 to 6 with an increment of 1 point.   We further conducted an ablation study to verify the effectiveness of each component in our model. We instructed the subjects to compare our proposed model to the models with one component ablated. The ablation study was conducted on OOD texts for more pronounced results. In addition, we train a logistic regression P2G predictor on the Wikipedia corpus to predict graphemes from phonemes to test whether the learned representation contains contextual grapheme information. 
 
\section{Results} 

\subsection{TTS Performance} As shown in Table \ref{tab:t1}, there is no significant improvement with PL-BERT on in-distribution texts. However, we can see that our model has significantly outperformed the baseline model where no pre-trained BERT is used in Table \ref{tab:t2} in terms of both MOS and CMOS for the out-of-distribution (OOD) texts. In particular, our model is significantly better than MP-BERT (Wilcoxon test, $p < 0.05$), with a CMOS of plus 0.16. This shows that training with phoneme predictions instead of sup-phoneme units makes the TTS model generalize better for unseen texts. We also notice a massive MOS gap between in-distribution texts and OOD texts. The MOS difference between StyleTTS w/ PL-BERT and ground truth is not statistically significant ($p$ > 0.05), although CMOS shows a slight preference of the raters for the ground truth over our model. However, the performance drops dramatically when the input texts are OOD. This performance gap is not specific to StyleTTS models; it is a prevalent problem for many TTS models, as shown in Table 2. These MOS scores are significantly worse than those reported in $\cite{li2022styletts}$ by roughly 0.4 to 0.5 points. The results suggest that future works should give more weight to evaluations on OOD texts. In addition, since our model does not need to process the sub-phoneme tokens, our model is 1.05 times faster than MX-BERT on a single NVIDIA A40 GPU. 

\subsection{Ablation Study} 
\label{sec:abla}
Table 3 shows a slight performance decrease when $\mathcal{L}_{P2G}$ is removed during training. However, the CMOS is still higher than the baseline StyleTTS model without BERT. This shows that using a pre-trained phoneme-level BERT improves downstream TTS tasks even when trained without $\mathcal{L}_{P2G}$. This can be attributed to our masking strategy where the entire grapheme phonemes are masked, so $\mathcal{L}_{MLM}$ alone can learn rich enough linguistic representation that helps the downstream TTS tasks. However, when $\mathcal{L}_{MLM}$ is removed, the CMOS drops dramatically, indicating that with only $\mathcal{L}_{P2G}$ the trained model cannot retain the input phonemes information and, therefore, cannot be used for downstream TTS tasks. We note that the P2G prediction accuracy decreases dramatically when $\mathcal{L}_{P2G}$ is removed from the training objectives. This shows that $\mathcal{L}_{MLM}$, even with whole-word masking, does not guarantee that the model learns word-level linguistic representations. This partly explains why our model performs better than MP-BERT, as MP-BERT lacks the  $\mathcal{L}_{P2G}$ objective that enables the model to learn linguistic representations at the phoneme level. 

\begin{table}[!t]
\caption{Ablation study for verifying the effectiveness of MLM, P2G, and BERT compared to StyleTTS w/ PL-BERT. \\}

\label{tab:4}
\small
\centering
\begin{tabular}{c|c|c|c}
\hline
Model & CMOS & ACC (Top-1) & ACC  (Top-5) \\ 
\hline
Proposed              & 0.00 & 67.48\%  & 90.33\% \\

 w/o $\mathcal{L}_{P2G}$   & -- 0.11 & 13.45 \% & 24.97\% \\
 w/o $\mathcal{L}_{MLM}$   & -- 4.57 & 68.73 \% & 90.24 \% \\
 w/o PL-BERT            & -- 0.44 & --- & --- \\
\hline
\end{tabular}
\end{table}

\section{Conclusions}
In this work, we proposed phoneme-level BERT, a phoneme-level language model that produces contextualized embeddings that improve the naturalness and prosody of downstream TTS tasks. Unlike previous works, our model takes only phonemes as input, greatly reducing the resources needed during training and inference. We show that our model has significantly outperformed the baseline StyleTTS model, where no BERT encoder is fine-tuned with the TTS model, and we also show that our pre-training strategy is better than MP-BERT for out-of-distribution (OOD) texts. We have identified a performance gap in many existing TTS models between in-distribution and OOD texts. Since in-distribution texts are barely used for real-world applications, we advocate that future studies focus more on TTS development for OOD texts.

\section{Acknowledgements}
We thank Gavin Mischler for providing feedback to the quality of models during the development stage of this work. This work was funded by the national institute of health (NIHNIDCD) and a grant from Marie-Josee and Henry R. Kravis.

\begingroup
\setstretch{0.9}
    \bibliographystyle{IEEEtran}
    \bibliography{mybib}

\begin{thebibliography}{10}
\providecommand{\url}[1]{#1}
\csname url@samestyle\endcsname
\providecommand{\newblock}{\relax}
\providecommand{\bibinfo}[2]{#2}
\providecommand{\BIBentrySTDinterwordspacing}{\spaceskip=0pt\relax}
\providecommand{\BIBentryALTinterwordstretchfactor}{4}
\providecommand{\BIBentryALTinterwordspacing}{\spaceskip=\fontdimen2\font plus
\BIBentryALTinterwordstretchfactor\fontdimen3\font minus
  \fontdimen4\font\relax}
\providecommand{\BIBforeignlanguage}[2]{{%
\expandafter\ifx\csname l@#1\endcsname\relax
\typeout{** WARNING: IEEEtran.bst: No hyphenation pattern has been}%
\typeout{** loaded for the language `#1'. Using the pattern for}%
\typeout{** the default language instead.}%
\else
\language=\csname l@#1\endcsname
\fi
#2}}
\providecommand{\BIBdecl}{\relax}
\BIBdecl

\bibitem{tan2022naturalspeech}
X.~Tan, J.~Chen, H.~Liu, J.~Cong, C.~Zhang, Y.~Liu, X.~Wang, Y.~Leng, Y.~Yi,
  L.~He \emph{et~al.}, ``Naturalspeech: End-to-end text to speech synthesis
  with human-level quality,'' \emph{arXiv preprint arXiv:2205.04421}, 2022.

\bibitem{tan2021survey}
X.~Tan, T.~Qin, F.~Soong, and T.-Y. Liu, ``A survey on neural speech
  synthesis,'' \emph{arXiv preprint arXiv:2106.15561}, 2021.

\bibitem{li2022styletts}
Y.~A. Li, C.~Han, and N.~Mesgarani, ``Styletts: A style-based generative model
  for natural and diverse text-to-speech synthesis,'' \emph{arXiv preprint
  arXiv:2205.15439}, 2022.

\bibitem{kenter2020improving}
T.~Kenter, M.~K. Sharma, and R.~Clark, ``Improving prosody of rnn-based english
  text-to-speech synthesis by incorporating a bert model,'' 2020.

\bibitem{xiao2020improving}
Y.~Xiao, L.~He, H.~Ming, and F.~K. Soong, ``Improving prosody with linguistic
  and bert derived features in multi-speaker based mandarin chinese neural
  tts,'' in \emph{ICASSP 2020-2020 IEEE International Conference on Acoustics,
  Speech and Signal Processing (ICASSP)}.\hskip 1em plus 0.5em minus
  0.4em\relax IEEE, 2020, pp. 6704--6708.

\bibitem{zhang2020unified}
Y.~Zhang, L.~Deng, and Y.~Wang, ``Unified mandarin tts front-end based on
  distilled bert model,'' \emph{arXiv preprint arXiv:2012.15404}, 2020.

\bibitem{xu2021improving}
G.~Xu, W.~Song, Z.~Zhang, C.~Zhang, X.~He, and B.~Zhou, ``Improving prosody
  modelling with cross-utterance bert embeddings for end-to-end speech
  synthesis,'' in \emph{ICASSP 2021-2021 IEEE International Conference on
  Acoustics, Speech and Signal Processing (ICASSP)}.\hskip 1em plus 0.5em minus
  0.4em\relax IEEE, 2021, pp. 6079--6083.

\bibitem{jia2021png}
Y.~Jia, H.~Zen, J.~Shen, Y.~Zhang, and Y.~Wu, ``Png bert: augmented bert on
  phonemes and graphemes for neural tts,'' \emph{arXiv preprint
  arXiv:2103.15060}, 2021.

\bibitem{zhang2022mixed}
G.~Zhang, K.~Song, X.~Tan, D.~Tan, Y.~Yan, Y.~Liu, G.~Wang, W.~Zhou, T.~Qin,
  T.~Lee \emph{et~al.}, ``Mixed-phoneme bert: Improving bert with mixed phoneme
  and sup-phoneme representations for text to speech,'' \emph{arXiv preprint
  arXiv:2203.17190}, 2022.

\bibitem{sennrich2015neural}
R.~Sennrich, B.~Haddow, and A.~Birch, ``Neural machine translation of rare
  words with subword units,'' \emph{arXiv preprint arXiv:1508.07909}, 2015.

\bibitem{cui2021pre}
Y.~Cui, W.~Che, T.~Liu, B.~Qin, and Z.~Yang, ``Pre-training with whole word
  masking for chinese bert,'' \emph{IEEE/ACM Transactions on Audio, Speech, and
  Language Processing}, vol.~29, pp. 3504--3514, 2021.

\bibitem{liu2019roberta}
Y.~Liu, M.~Ott, N.~Goyal, J.~Du, M.~Joshi, D.~Chen, O.~Levy, M.~Lewis,
  L.~Zettlemoyer, and V.~Stoyanov, ``Roberta: A robustly optimized bert
  pretraining approach,'' \emph{arXiv preprint arXiv:1907.11692}, 2019.

\bibitem{kuchaiev2019nemo}
O.~Kuchaiev, J.~Li, H.~Nguyen, O.~Hrinchuk, R.~Leary, B.~Ginsburg, S.~Kriman,
  S.~Beliaev, V.~Lavrukhin, J.~Cook \emph{et~al.}, ``Nemo: a toolkit for
  building ai applications using neural modules,'' \emph{arXiv preprint
  arXiv:1909.09577}, 2019.

\bibitem{Bernard2021}
\BIBentryALTinterwordspacing
M.~Bernard and H.~Titeux, ``Phonemizer: Text to phones transcription for
  multiple languages in python,'' \emph{Journal of Open Source Software},
  vol.~6, no.~68, p. 3958, 2021. [Online]. Available:
  \url{https://doi.org/10.21105/joss.03958}
\BIBentrySTDinterwordspacing

\bibitem{ljspeech17}
K.~Ito and L.~Johnson, ``The lj speech dataset,''
  \url{https://keithito.com/LJ-Speech-Dataset/}, 2017.

\bibitem{kong2020hifi}
J.~Kong, J.~Kim, and J.~Bae, ``Hifi-gan: Generative adversarial networks for
  efficient and high fidelity speech synthesis,'' \emph{Advances in Neural
  Information Processing Systems}, vol.~33, 2020.

\bibitem{lan2019albert}
Z.~Lan, M.~Chen, S.~Goodman, K.~Gimpel, P.~Sharma, and R.~Soricut, ``Albert: A
  lite bert for self-supervised learning of language representations,''
  \emph{arXiv preprint arXiv:1909.11942}, 2019.

\bibitem{gerlach2020standardized}
M.~Gerlach and F.~Font-Clos, ``A standardized project gutenberg corpus for
  statistical analysis of natural language and quantitative linguistics,''
  \emph{Entropy}, vol.~22, no.~1, p. 126, 2020.

\bibitem{reimers-2020-multilingual-sentence-bert}
\BIBentryALTinterwordspacing
N.~Reimers and I.~Gurevych, ``Making monolingual sentence embeddings
  multilingual using knowledge distillation,'' \emph{arXiv preprint
  arXiv:2004.09813}, 04 2020. [Online]. Available:
  \url{http://arxiv.org/abs/2004.09813}
\BIBentrySTDinterwordspacing

\bibitem{li2021starganv2}
Y.~A. Li, A.~Zare, and N.~Mesgarani, ``Starganv2-vc: A diverse, unsupervised,
  non-parallel framework for natural-sounding voice conversion,'' \emph{arXiv
  preprint arXiv:2107.10394}, 2021.

\bibitem{li2022stylettsvc}
Y.~A. Li, C.~Han, and N.~Mesgarani, ``Styletts-vc: One-shot voice conversion by
  knowledge transfer from style-based tts models,'' \emph{arXiv preprint
  arXiv:2212.14227}, 2022.

\bibitem{shen2018natural}
J.~Shen, R.~Pang, R.~J. Weiss, M.~Schuster, N.~Jaitly, Z.~Yang, Z.~Chen,
  Y.~Zhang, Y.~Wang, R.~Skerrv-Ryan \emph{et~al.}, ``Natural tts synthesis by
  conditioning wavenet on mel spectrogram predictions,'' in \emph{2018 IEEE
  International Conference on Acoustics, Speech and Signal Processing
  (ICASSP)}.\hskip 1em plus 0.5em minus 0.4em\relax IEEE, 2018, pp. 4779--4783.

\bibitem{ren2021fastspeech}
\BIBentryALTinterwordspacing
Y.~Ren, C.~Hu, X.~Tan, T.~Qin, S.~Zhao, Z.~Zhao, and T.-Y. Liu, ``Fastspeech 2:
  Fast and high-quality end-to-end text to speech,'' in \emph{International
  Conference on Learning Representations}, 2021. [Online]. Available:
  \url{https://openreview.net/forum?id=piLPYqxtWuA}
\BIBentrySTDinterwordspacing

\bibitem{kim2021conditional}
J.~Kim, J.~Kong, and J.~Son, ``Conditional variational autoencoder with
  adversarial learning for end-to-end text-to-speech,'' in \emph{Proceedings of
  the 38th International Conference on Machine Learning}, ser. Proceedings of
  Machine Learning Research, M.~Meila and T.~Zhang, Eds., vol. 139, 18--24 Jul
  2021, pp. 5530--5540.

\end{thebibliography}
\endgroup

\end{document}